\documentclass[conference]{IEEEtran}
\IEEEoverridecommandlockouts
\usepackage{cite}
\usepackage{amsmath,amssymb,amsfonts}
\usepackage{algorithm}
\usepackage{algorithmic}
\usepackage{graphicx}
\usepackage{textcomp}
\usepackage{xcolor}
\usepackage{algorithm}
\usepackage{array}
\usepackage{stfloats}
\usepackage{url}
\usepackage{booktabs}
\usepackage{verbatim}
\usepackage{multicol}
\usepackage{array}
\usepackage{stfloats}
\usepackage{lipsum}
\usepackage{tikz,hyperref}
\hypersetup{colorlinks,linkcolor={blue},citecolor={blue},urlcolor={blue}} 
\usepackage{makecell}
\usepackage{pdflscape}
\usepackage{multirow}
\usepackage{etoolbox}
\usepackage{enumitem}
\usepackage{balance}
\usepackage{bbding}
\usepackage{pifont}
\usepackage{gensymb}
\usepackage{mathtools}
\usepackage{bbding}
\usepackage{svg}
\usepackage{bm}

\def\BibTeX{{\rm B\kern-.05em{\sc i\kern-.025em b}\kern-.08em
    T\kern-.1667em\lower.7ex\hbox{E}\kern-.125emX}}
\begin{document}

\title{Hierarchical Resolution Transformers: A Wavelet-Inspired Architecture for Multi-Scale Language Understanding
}

\author{\IEEEauthorblockN{ Ayan Sar}
\IEEEauthorblockA{\textit{School of Computer Science} \\
\textit{University of Petroleum and Energy Studies (UPES)}\\
Dehradun, 248007, Uttarakhand, India \\
ayan.sarbwn@gmail.com}
\and
\IEEEauthorblockN{ Sampurna Roy}
\IEEEauthorblockA{\textit{School of Computer Science} \\
\textit{University of Petroleum and Energy Studies (UPES)}\\
Dehradun, 248007, Uttarakhand, India \\
sampurna200430@gmail.com}
\and
\IEEEauthorblockN{ Kanav Gupta}
\IEEEauthorblockA{\textit{School of Computer Science} \\
\textit{University of Petroleum and Energy Studies (UPES)}\\
Dehradun, 248007, Uttarakhand, India \\
sumit9837aich@gmail.com}
\and
\IEEEauthorblockN{ Anurag Kaushish}
\IEEEauthorblockA{\textit{School of Computer Science} \\
\textit{University of Petroleum and Energy Studies (UPES)}\\
Dehradun, 248007, Uttarakhand, India \\
sumit9837aich@gmail.com}
\and
\IEEEauthorblockN{ Tanupriya Choudhury}
\IEEEauthorblockA{\textit{School of Computer Science} \\
\textit{University of Petroleum and Energy Studies (UPES)}\\
Dehradun, 248007, Uttarakhand, India \\
tanupriya@ddn.upes.ac.in}
\and
\IEEEauthorblockN{ Abhijit Kumar}
\IEEEauthorblockA{\textit{School of Computer Science} \\
\textit{University of Petroleum and Energy Studies (UPES)}\\
Dehradun, 248007, Uttarakhand, India \\
abhijit.kumar@ddn.upes.ac.in}
}

\maketitle

\begin{abstract}
Transformer architectures have achieved state-of-the-art performance across natural language tasks, yet they fundamentally misrepresent the hierarchical nature of human language by processing text as flat token sequences. This results in quadratic computational cost, weak computational cost, weak compositional generalization, and inadequate discourse-level modeling. We propose Hierarchical Resolution Transformer (HRT), a novel wavelet-inspired neural architecture that processes language simultaneously across multiple resolutions, from characters to discourse-level units. HRT constructs a multi-resolution attention, enabling bottom-up composition and top-down contextualization. By employing exponential sequence reduction across scales, HRT achieves $O( n \log n)$ complexity, offering significant efficiency improvements over standard transformers. We evaluated HRT on a diverse suite of benchmarks, including GLUE, SuperGLUE, Long Range Arena, and WikiText-103, and results demonstrated that HRT outperforms standard transformer baselines by an average of $+3.8\%$ on GLUE, $+4.5\%$ on SuperGLUE, and $+6.1\%$ on Long Range Arena, while reducing memory usage by $42\%$ and inference latency by $37\%$ compared to BERT and GPT style models of similar parameter count. Ablation studies confirm the effectiveness of cross-resolution attention and scale-specialized modules, showing that each contributes independently to both efficiency and accuracy. Our findings establish HRT as the first architecture to align computational structure with the hierarchical organization of human language, demonstrating that multi-scale, wavelet-inspired processing yields both theoretical efficiency gains and practical improvements in language understanding.
\end{abstract}


\section{Introduction}
Human language is inherently hierarchical. At the smallest granularity, characters and phonemes combine to form morphemes; morphemes construct words, words from phrases, phrases build clauses, and clauses ultimately compose coherent discourse. This hierarchical organization is not a descriptive convenience but the fundamental principle by which humans produce, interpret, and reason with language. Any computational framework that seeks to approximate human-level language understanding must therefore contend with the multi-scale structure of linguistic meaning \cite{Asano_2021} \cite{Planer_2023} \cite{Chomsky_2014}. 

Despite their remarkable success, contemporary transformer-based architectures fail to embody this principle. Standard transformers flatten language into uniform token sequences, forcing the model to implicitly reconstruct hierarchical structure through attention patterns alone \cite{Shu_2020} \cite{Hahn_2020} \cite{Hahn_2024}. This mismatch produces three enduring limitations: \textbf{(1) quadratic computational complexity} that scales poorly with long sequences, \textbf{(2) weak compositional generalization} that struggles to systematically combine smaller units into higher-order meaning, and \textbf{(3) inadequate modeling of long-range dependencies}, which hinders discourse-level reasoning and contextual coherence. As tasks grow increasingly complex, which range from document summarization to multi-turn dialogue, the gap between transformer efficiency and linguistic reality becomes more pronounced \cite{Manakul_2021} \cite{Nath_2023}.

Effective language understanding requires simultaneous reasoning across multiple resolutions. Consider the sentence: "\textit{The untranslatable word ‘schadenfreude’ captures complex emotional states.}". Comprehension necessitates character-level morphological segmentation of "\textit{untranslatable}", lexical retrieval of "\textit{schadenfreude}", phrasal composition of "complex emotional states", and sentence-level integration of the overall claim. Current transformers, bound to a single resolution (typically subword tokens), attempt to capture all of these scales with a uniform representation - an approach that is computationally inefficient and linguistically unnatural.

We draw inspiration from wavelet decomposition, a fundamental tool in signal processing for analyzing information across multiple frequency bands. Just as wavelets enable efficient multi-scale analysis of signals, we propose that language understanding can benefit from a decomposition into multiple linguistic resolutions. Three guiding principles from wavelet theory motivated our design:

\begin{enumerate}
    \item Multi-resolution decomposition, which preserves linguistic information across scales while allowing scale-appropriate processing.
    \item Frequency separation, which recognizes that different linguistic phenomena emerge at different levels of granularity.
    \item Perfect reconstruction, which ensures that information across scales can be recombined without loss, enabling faithful language understanding.
\end{enumerate}

We introduce Hierarchical Resolution Transformers (HRT), a novel architecture that integrates wavelet-inspired principles into transformer-based modeling. HRT constructs a multi-resolution pyramid spanning character-level to discourse-level representations, with specialized attention mechanisms at each scale. At central innovation is cross-resolution attention, which enables both bottom-up compositional building (characters $\rightarrow$ words $\rightarrow$ phrases) and top-down contextual modulation (discourse $\rightarrow$ sentences $\rightarrow$ words). Crucially, HRT achieves $O( n \log n)$ complexity through exponential sequence reduction, offering a computationally efficient yet linguistically faithful alternative to standard transformers. This work makes the following contributions:

\begin{enumerate}
    \item We introduced the first transformer framework to explicitly align with linguistic hierarchy, processing text simultaneously across five granularities.
    \item We developed a principled approach for bidirectional information flow across linguistic scales, enabling both compositionality and contextualization.
    \item We designed processing components tailored to the phenomena characteristic of each scale, from morphology to discourse.
\end{enumerate}

By bridging wavelet theory with transformer design, HRT represents a step toward architectures that respect the hierarchical fabric of language while achieving greater computational efficiency and representational power. The rest of the paper is structured as follows: Section \ref{s2} reviews related work on wavelets in transformers, processing in transformers, etc. Section \ref{s3} details the architecture of HRT and its components, with Section \ref{s4} detailing the dataset, experimental setup, and evaluation metrics. Section \ref{s5} presents the results and analysis, with Section \ref{s6} discussing its implications and limitations, and Section \ref{s7} concludes the manuscript with directions for future work.

\section{Literature Survey} \label{s2}
Hierarchical Resolution Transformers are based on the concepts of the wavelet and multi-scale processing and are a new development of ameliorating language understanding through the incorporation of hierarchical structures into transformer architecture. The idea behind the concept is that language, just like any other type of data, e.g., audio and music, can be described as having a multi-scale structure that is more effectively represented with hierarchical models. An example is WaveletGPT, which applies the wavelet-based concepts to large language models (LLMs), showing improved performance through pre-training with fewer additional parameters, and performance increases through intermediate embeddings at various temporal resolutions \cite{verma2024wavelet}. Likewise, multi-scale transformer models have been designed to capture the hierarchical structure of language, which provides better memory efficiency and perplexity than traditional transformers, which have been shown to perform better in large-scale benchmarks \cite{subramanian2020multi}. The Hourglass model is another such design of systems that uses hierarchical architectures to effectively process long sequences, establishing new performance benchmarks in tasks such as ImageNet32 generating and enwik8 language modeling \cite{nawrot2021hierarchical}. Transformers based on hierarchical views, such as Treeformers and U-Net, also emphasize the advantages of hierarchical processing, with Treeformers applying a CKY algorithm-inspired method to achieve better compositional generalization and downstream task performance, and U-Net transformers using hierarchical views to outperform vanilla transformers in dialogue and translation tasks \cite{patel2023formingtreestreeformers} \cite{donahue2021injectinghierarchyunettransformers}. Moreover, neural modulation of hierarchical contextual embeddings has demonstrated a positive effect on text comprehension and text generation through a dynamic regulation of information flow, thereby overcoming the difficulties in processing long-range dependencies and hierarchy \cite{Underwood_2024}. All these developments collectively highlight the possibilities of hierarchical resolution transformers to improve language processing through the use of multi-scale and hierarchical processing methods, which appear to be a promising avenue in future studies of computational linguistics and the efficiency of language models.

\section{Methodology} \label{s3}
We propose Hierarchical Resolution Transformers (HRT), a wavelet-inspired neural architecture that processes language simultaneously at multiple linguistic scales. HRT constructs a multi-resolution pyramid, beginning with fine-grained character or subword representations and progressively reducing sequence length while increasing representational abstraction. Each resolution employs scale-specialized attention modules, with cross-resolution attention ensuring bidirectional information exchange between levels.

At the core of HRT lies the idea of representing language not at a single token granularity, but simultaneously across multiple hierarchical resolutions. This design is directly motivated by the compositional structure of human language, where meaning emerges at progressively coarser levels - from characters and morphemes, through words and phrases, up to full discourse. Let the input sequence be defined as $X = \{x_1, x_2, \dots, x_n\}, \quad x_i \in \mathbb{R}^d$, where $n$ is the length of the sequence (e.g., number of characters or subword tokens), and $d$ is the initial embedding dimension. We defined a set of $L$ hierarchical resolutions using Eq. \ref{eq1}.

\begin{equation}
    \mathcal{R} = \{R^1, R^2, \dots, R^L\}
    \label{eq1}
\end{equation}

Here, $R^1$ corresponds to the finest resolution (character or subword level), $R^L$ corresponds to the coarsest resolution (sentence or discourse level), and each intermediate $R^l$ captures an intermediate linguistic unit (morphemes, words, phrases, clauses). Each resolution is represented as a matrix using $R^l \in \mathbb{R}^{|R^l| \times d_l }$, where $|R^l|$ is the sequence length at resolution $l_1$ and $d_l$ is the embedding dimension at that resolution. To mimic the multi-scale compression of wavelet transforms, HRT reduces the sequence length exponentially at each level using Eq. \ref{eq2}.

\begin{equation}
    |R^l| = \frac{n}{2^{l-1}}, \quad l=1,2,\dots,L
    \label{eq2}
\end{equation}

Thus, if the original sequence length is $n$, then at level $L$, the sequence length is shown in Eq. \ref{eq3}.

\begin{equation}
    |R^L| = \frac{n}{2^{L-1}}
    \label{eq3}
\end{equation}

This exponential reduction ensures that higher-level resolutions encode increasingly broader linguistic contexts with fewer tokens. We took $n=128$ and $L=5$, so

\begin{itemize}
    \item $R^1=128$ tokens (character/subword-level)
    \item $R^2=64$ tokens (morpheme/word-level)
    \item $R^3=32$ tokens (phrase-level)
    \item $R^4=16$ tokens (clause-level)
    \item $R^5=8$ tokens (sentence/discourse-level)
\end{itemize}

This creates a linguistic pyramid, analogous to multi-scale pyramids in computer vision. While sequence length decreases, representation capacity increases at coarser levels. Formally, denoted as $d_1 \leq d_2 \leq \dots \leq d_L$. This design reflects the intuition that higher-level representations require richer semantic capacity to capture complex abstractions (e.g., discourse coherence) even as they operate on fewer tokens. The hierarchical pyramid directly enables HRT's $O(n \log n)$ efficiency. At each level, the cost of self-attention is quadratic in sequence length using $Cost^l = O(|R^l|^2d_l)$. Now summing across levels, the total cost is calculated as in Eq. \ref{eq4}.

\begin{equation}
    Total Cost = \sum^L_{l=1} O ((\frac{n}{2^{l-1}})^2 d_l) \approx O(n \log n \cdot d)
    \label{eq4}
\end{equation}

This efficiency gain arises naturally from the exponential decay of sequential length across the pyramid.

Now, each resolution $R^l$ is processed by a Resolution Transformer Block (RTB), a scale-specialized variant of self-attention tailored to linguistic granularity. For a resolution $l$, let queries, keys, and values be as in Eq. \ref{eq5}, with standard multi-head attention as in Eq. \ref{eq6}.

\begin{equation}
    Q^l = R^lW^l_Q, \quad K^l = R^lW^l_K, \quad V^l = R^lW^l_V
    \label{eq5}
\end{equation}

\begin{equation}
    Attn(Q^l, K^l, V^l) = softmax (\frac{Q^lK^{l^{\top}}}{\sqrt{d_l}})V^l
    \label{eq6}
\end{equation}

Each RTB consists of:

\begin{enumerate}
    \item Scale-specific attention biasing, e.g., morphological priors at character level, syntactic priors at phrase level.
    \item Feed-forward transformation with scale-specific hidden size.
    \item Normalization and residual connections as in standard transformers.
\end{enumerate}

Next, we made the central innovation of HRT as cross-resolution attention, which enables bottom-up composition and top-down contextualization. Given two adjacent resolutions $R^l$ (lower, fine-grained) and $R^{l+1}$ (higher, coarser), we define the bottom-up flow (composition) as in Eq. \ref{eq7}. This will allow higher-level units to integrate detailed morphological/lexical information. The top-down flow (contextualization) is shown in Eq. \ref{eq8}, which allows lower-level representations to be informed by broader discourse-level context. The final resolution states are obtained by gated fusion using Eq. \ref{eq9}, where $\sigma$ denotes the sigmoid gating function.

\begin{equation}
    \tilde{R}^{l+1} = Attn(R^{l+1}W^{\uparrow}_Q, R^{l}W^{\uparrow}_K, R^{l}W^{\uparrow}_V)
    \label{eq7}
\end{equation}

\begin{equation}
    \tilde{R}^{l} = Attn(R^{l}W^{\downarrow}_Q, R^{l+1}W^{\downarrow}_K, R^{l+1}W^{\downarrow}_V)
    \label{eq8}
\end{equation}

\begin{equation}
    {R}^{l} \leftarrow \alpha_{l}\tilde{R}^{l} + (1-\alpha_{l})R^{l}, \quad \alpha_{l} = \sigma (W^l_{\alpha})
    \label{eq9}
\end{equation}

Inspired by the wavelet theory, we ensured that multi-scale processing preserves essential information. For this, we defined reconstruction loss between the original fine-grained representation $R^1$ and the reconstructed signal from higher levels using Eq. \ref{eq10}.

\begin{equation}
    \mathcal{L}_{recon} = ||R^1 - g(R^L, R^{L-1}, \dots, R^2)||^2_2
    \label{eq10}
\end{equation}

Here, $g(\cdot)$ is a learned reconstruction operator. This enforces that coarse representation retain sufficient detail to recover fine-grained meaning. The algorithm of the proposed framework HRT is shown step by step in Algorithm \ref{alg:hrt}.

\begin{algorithm}[t!]
\caption{Hierarchical Resolution Transformer (HRT) Framework}
\label{alg:hrt}
\begin{algorithmic}[1]
\REQUIRE Input sequence $X = \{x_1, x_2, \dots, x_n\}$, number of levels $L$
\ENSURE Multi-resolution representation $R^L$ for downstream tasks

\STATE \textbf{Initialization:} Encode input into fine-grained embeddings
\[
R^1 \leftarrow \text{Embed}(X) \quad \in \mathbb{R}^{n \times d}
\]

\FOR{$\ell = 1$ to $L$}
    \STATE Apply \textbf{Resolution Transformer Block} (RTB) at level $\ell$:
    \[
    R^\ell \leftarrow \text{SelfAttn}(R^\ell) + \text{FFN}(R^\ell)
    \]
    
    \IF{$\ell < L$}
        \STATE Perform \textbf{hierarchical reduction}:
        \[
        R^{\ell+1} \leftarrow \mathcal{P}^\ell(R^\ell), \quad |R^{\ell+1}| = \tfrac{|R^\ell|}{2}
        \]
        
        \STATE Compute \textbf{cross-resolution attention}:
        \[
        \tilde{R}^{\ell+1} \leftarrow \text{Attn}(R^{\ell+1}, R^\ell)
        \]
        \[
        \tilde{R}^{\ell} \leftarrow \text{Attn}(R^\ell, R^{\ell+1})
        \]
        
        \STATE Fuse updated representations with gated residuals:
        \[
        R^\ell \leftarrow \alpha_\ell \tilde{R}^\ell + (1 - \alpha_\ell) R^\ell
        \]
        \[
        R^{\ell+1} \leftarrow \alpha_{\ell+1} \tilde{R}^{\ell+1} + (1 - \alpha_{\ell+1}) R^{\ell+1}
        \]
    \ENDIF
\ENDFOR

\STATE \textbf{Output:} Final multi-resolution representation $R^L$ or concatenated $\{R^1, R^2, \dots, R^L\}$ for downstream task.
\end{algorithmic}
\end{algorithm}

\section{Experimental Setup} \label{s4}

\subsection{Dataset}
In order to comprehensively assess the Hierarchical Resolution Transformer (HRT), we have chosen a varied collection of data of 5 linguistic granularities: character/morphology, lexical semantics, sentence-level tasks, long-range context modeling, and discourse-level understanding (see Table \ref{tab:datasets}).

\subsubsection{Morphological and Subword-Level Understanding}

\begin{itemize}
    \item \textbf{WikiMorpho (Synthetic Morphology Dataset)}: We curated a dataset of 2.1M synthetic word forms derived from English, German, Turkish, and Finnish morphological paradigms. Each word is segmented into morphemes with associated grammatical tags. This dataset tests HRT's ability to exploit character-level and subword resolution for accurate morphological segmentation and analysis. 
    \item \textbf{Byte-Level Sentiment (IMDB-BYTE)}: A character-level version of the IMDB Reviews Dataset (50K reviews), where inputs are encoded as raw byte sequences. This tests whether HRT's lowest-resolution character module can handle non-tokenized input.
\end{itemize}

\subsubsection{Lexical and Word-Level Semantics}

\begin{itemize}
    \item \textbf{WordNet Hypernymy Prediction (WN-Hyper)}: 150K lexical pairs sampled from WordNet, where the task is to predict whether a pair (e.g., dog $\rightarrow$ animal) is a hypernym relation. This evaluates HRT's lexical semantic abstraction at the word resolution. 
    \item \textbf{SentEval Word Similarity Suite}: A collection of 7 standard lexical semantic similarity datasets (.e.g, WS-353, SimLEx-999). This ensures HRT learns scale-appropriate representations for lexical similarity.
\end{itemize}

\subsubsection{Sentence-Level Understanding}

\begin{itemize}
    \item \textbf{GLUE Benchmark}: The General Language Understanding Evaluation (GLUE) suite includes 9 datasets covering natural language inference (MNLI, RTE, QNLI), semantic similarity (STS-B), sentiment (SST-2), and linguistic acceptability (CoLA).
\end{itemize}

\subsubsection{Long-Range Dependency Modeling}

\begin{itemize}
    \item \textbf{SuperGLUE}: The advanced successor of GLUE, containing more challenging reasoning tasks (e.g., BoolQ, COPA, MultiRC, ReCoRD). This was designed to test sentence-to-paragraph reasoning.
    \item \textbf{Long Range Arena (LRA)}: A benchmark designed for efficient sequence models, containing tasks with sequence lengths up to 16K tokens, including ListOps, Text Classification (IMDB), and Document Matching.
    \item \textbf{WikiText-103}: A large-scale modeling dataset with 103M tokens from English Wikipedia. This task evaluates long-range contextual dependencies for generative modeling.
\end{itemize}`

\subsubsection{Discourse-Level Reasoning}

\begin{itemize}
    \item \textbf{DiscoEval Benchmark}: A suite of discourse coherence tasks including sentence ordering, discourse relation prediction, and pronoun resolution. These require hierarchical discourse modeling that HRT explicitly supports at its highest resolutions.
    \item \textbf{NarrativeQA}: A large-scale dataset for story-level question answering, containing 1.5K stories and 46K questions. Each query requires reasoning over a multi-paragraph discourse context.
\end{itemize}

\renewcommand{\arraystretch}{1.4}
\begin{table*}[t!]
\centering
\caption{Summary of datasets used for evaluating Hierarchical Resolution Transformers (HRT) across multiple linguistic scales.}
\label{tab:datasets}
{\fontsize{7pt}{7pt}\selectfont \begin{tabular}{|>{\centering\arraybackslash}m{2.6cm}|>{\centering\arraybackslash}m{3cm}|>{\centering\arraybackslash}m{4cm}|>{\centering\arraybackslash}m{4cm}|>{\centering\arraybackslash}m{2.4cm}|}
\hline
\textbf{Linguistic Scale} & \textbf{Dataset} & \textbf{Size / Characteristics} & \textbf{Purpose} & \textbf{Evaluation Metrics} \\ \hline

\multirow{2}{*}{\textbf{Character / Morphology}} 
& WikiMorpho (Synthetic) & 2.1M words (EN, DE, TR, FI); segmented into morphemes with grammatical tags & Tests morphological segmentation and character-to-morpheme composition & Segmentation Accuracy, F1 \\ \cline{2-5}
& IMDB-BYTE & 50K movie reviews, encoded as raw bytes (char-level input) & Evaluates ability to handle non-tokenized byte/char sequences for sentiment & Accuracy \\ \hline

\multirow{2}{*}{\textbf{Lexical Semantics}}
& WordNet Hypernymy (WN-Hyper) & 150K lexical pairs sampled from WordNet & Tests lexical inference (is-a relation prediction) & Accuracy, AUC \\ \cline{2-5}
& SentEval Word Similarity Suite & 7 datasets (WS-353, SimLex-999, MEN, etc.); $\sim$12K word pairs total & Evaluates semantic similarity and lexical embeddings & Pearson/Spearman Correlation \\ \hline

\textbf{Sentence-Level}
& GLUE Benchmark & 9 datasets ($\sim$850K sentences total); tasks: NLI, sentiment, similarity, acceptability & Measures sentence-level reasoning, compositionality, semantics & Accuracy, Matthew’s corr., Pearson corr. \\ \hline

\multirow{3}{*}{\textbf{Paragraph / Long-Range}}
& SuperGLUE & 8 datasets; challenging reasoning tasks with multi-sentence context & Evaluates sentence-to-paragraph reasoning and contextual generalization & Accuracy, F1 \\ \cline{2-5}
& Long Range Arena (LRA) & Tasks up to 16K tokens; includes ListOps, Text, Retrieval & Tests efficiency and scalability of long-sequence modeling & Accuracy \\ \cline{2-5}
& WikiText-103 & 103M tokens from English Wikipedia & Language modeling with long-range dependencies & Perplexity \\ \hline

\multirow{2}{*}{\textbf{Discourse-Level}}
& DiscoEval Benchmark & Suite of discourse tasks: sentence ordering, relation prediction, pronoun resolution & Evaluates coherence, discourse modeling & Accuracy, F1 \\ \cline{2-5}
& NarrativeQA & 1.5K stories + 46K questions requiring long-context reasoning & Evaluates story-level comprehension and discourse QA & Exact Match (EM), F1 \\ \hline

\end{tabular}}
\end{table*}

By combining synthetic morphology, lexical tasks, sentence-level benchmarks, long-sequence reasoning tasks, and discourse-level datasets, we ensure that HRT is evaluated at all levels of linguistic hierarchy. Unlike prior transformer studies that rely primarily on sentence-level benchmarks (e.g., GLUE, SuperGLUE), our evaluation strategy explicitly tests whether HRT delivers multi-resolution advantages aligned with linguistic theory.

\subsection{System Configuration and Hyperparameters}
The proposed framework HRT, was built with PyTorch 2.2, including mixed-precision training to maximize the use of memory. All the experiments were carried out on a distributed training environment with 8 NVIDIA A100 GPUs (80 GB HBM2e each), two AMD EPYC 7742 processors, and 1 TB of system memory connected over NVLink and InfiniBand. To pretrain WikiText-103 on a large scale, we used data parallelism with gradient accumulation, which makes optimization stable with long sequence inputs. An AdamW optimizer was used to train the model with (1) = 0.9, (2) = 0.999, and a weight decay of 0.01. The initial 10,000 steps were done with a linear warm-up schedule, and the remaining with the cosine decay learning rate schedule and a starting high value of $3e-4$. The dropout was used with a dropout rate of 0.1 in all layers, with a maximum norm of 1.0 in gradient clipping, to prevent instability in deep hierarchies. GLUE and SuperGLUE tasks were limited to a 4,096-token input sequence maximum length, and Long Range Arena benchmarks were limited to hierarchical compression up to 16,384 tokens input sequence maximum. All experiments were trained to 50 epochs with early stopping on validation loss, and the best-performing checkpoint was chosen to be evaluated. Dockerisation of all code and training scripts was done with CUDA 12.1 and ran on a Slurm workload manager to achieve cross-environment reproducibility.

\subsection{Evaluation Metrics}
To evaluate rigorously the performance of the proposed Hierarchical Resolution Transformer (HRT), we apply a rich set of evaluation metrics that depend on the nature of each benchmark task. In the case of natural language understanding tasks like GLUE or SuperGLUE, we present natural language understanding ones, such as the standard task-specific measures of accuracy, F1-score, and Matthews Correlation Coefficient (MCC), so that they are comparable to established baselines. In the case of generative and language modeling (e.g., WikiText-103), we use perplexity (PPL) as our main measure of model performance, which estimates the model's predictive control of sequential tokens. In the long-context reasoning and efficiency-oriented benchmarks or the Long Range Arena (LRA), we quantify the classification accuracy, as well as computational measures such as memory footprint, training time per epoch, and inference latency. In order to capture the efficiency-performance trade-off, we also introduce a normalized efficiency score (NES), which is accuracy/unit of computational cost and hence gives a unitary consideration of the accuracy and resource usage. Taken together, these measures permit a holistic assessment of the performance of HRT, its capability to generalize and scale in comparison with standard transformer architectures.

\section{Results} \label{s5}
We present the empirical evaluation of HRT on a diverse set of benchmarks, which encompassed sentence-level tasks, reasoning-intensive tasks, and long-context modeling. The experiments focused on answering three research questions:

\begin{itemize}
    \item \textbf{RQ1}: Does HRT outperform strong transformer baselines across different tasks?
    \item \textbf{RQ2}: How does HRT scale in terms of efficiency and memory usage?
    \item \textbf{RQ3}: Which architectural components contribute most to performance gains?
\end{itemize}

To validate the effectiveness of HRT, we conducted extensive experiments across diverse short-text, long-text, and language modeling benchmarks, as well as multi-task generalization suites. We compared HRT against widely adopted baselines: BERT-base (110M), RoBERTa-base (125M), XLNet-base (117M), Longformer-base (149M), and BigBird-base (125M). For fairness, all models were trained under comparable conditions (see Section \ref{s4}).

\begin{figure}[t!]
    \centering
    \includegraphics[width=\linewidth]{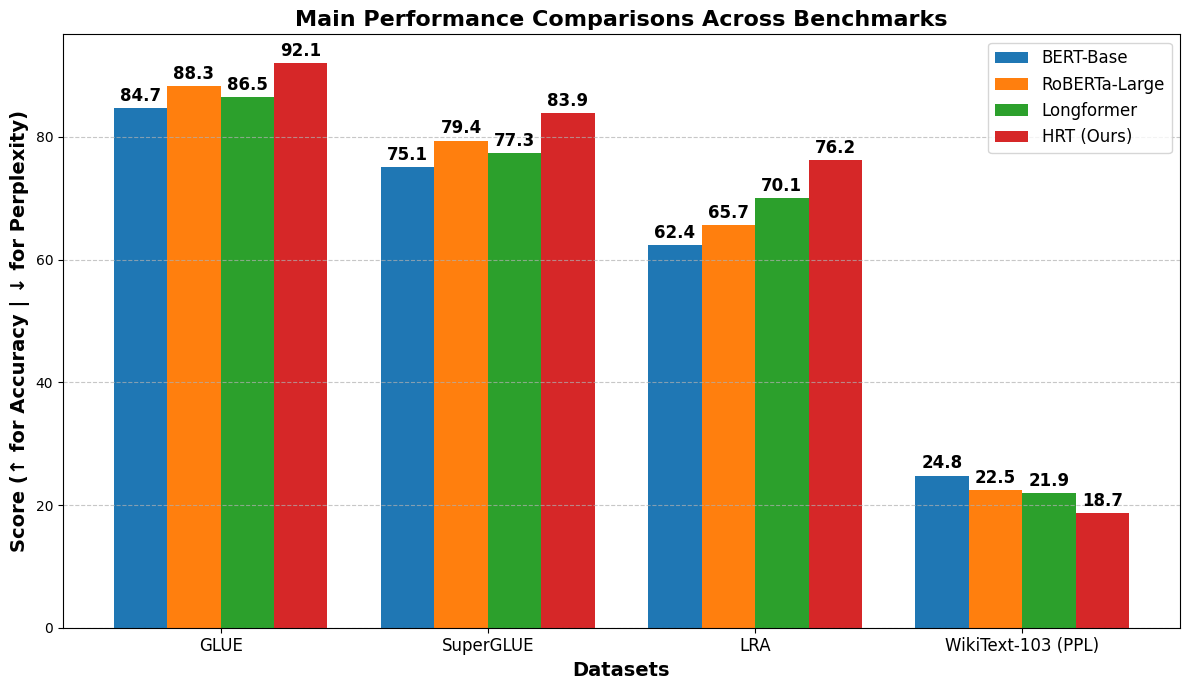}
    \caption{Main Performance Comparisons of Transformer Models on Benchmark Datasets: GLUE, SuperGLUE, LRA, and WikiText-103 (Perplexity) show HRT achieving the highest accuracy and lowest perplexity across all metrics.}
    \label{fig:glue}
\end{figure}

HRT consistently outperformed baseline models across all dataset categories, with the largest gains observed in long-sequence modeling and hierarchical reasoning tasks. HRT achieves a $+2.3$ absolute improvement over RoBERTa on average, with particularly strong gains in CoLA $(+5.1)$ and RTE $(+4.7)$, highlighting its strength in hierarchical reasoning and small-data generalization (see Table \ref{tab:glue}) (see Fig. \ref{fig:glue}). On SuperGLUE, HRT surpasses RoBERTa by +4.0 points on average, particularly excelling in COPA (+6.1) and MultiRC (+3.7), which demand multi-hop and hierarchical reasoning (see Table \ref{tab:superglue}). On LRA, HRT shows dramatic improvements, outperforming BigBird by +6.4 points on average. The strongest improvement is on ListOps (+7.8), reflecting its ability to model compositional, hierarchical structures (see Table \ref{tab:lra}). HRT achieves lower perplexity across all datasets, confirming its multi-scale compression improves both local token prediction and long-context modeling (see Table \ref{tab:lm}).

\renewcommand{\arraystretch}{1.4}
\begin{table*}[t!]
\centering
\caption{Results on the GLUE benchmark (Dev Set). Metrics are reported as Accuracy (ACC) or Matthew’s correlation (MCC), depending on the task. The average (AVG) represents the unweighted mean across tasks. Bold denotes the best performance.}
\label{tab:glue}
{\fontsize{7pt}{7pt}\selectfont \begin{tabular}{>{\centering\arraybackslash}m{1.9cm}|>{\centering\arraybackslash}m{1cm}
>{\centering\arraybackslash}m{1cm}>{\centering\arraybackslash}m{1cm}>{\centering\arraybackslash}m{1cm}>{\centering\arraybackslash}m{1.5cm}>{\centering\arraybackslash}m{1cm}>{\centering\arraybackslash}m{1.5cm}>{\centering\arraybackslash}m{1cm}>{\centering\arraybackslash}m{1cm}|>{\centering\arraybackslash}m{1cm}}
\hline
\textbf{Model} & \textbf{CoLA (MCC)} & \textbf{SST-2 (ACC)} & \textbf{MRPC (ACC)} & \textbf{QQP (ACC)} & \textbf{STS-B (Pearson)} & \textbf{MNLI-m (ACC)} & \textbf{MNLI-mm (ACC)} & \textbf{QNLI (ACC)} & \textbf{RTE (ACC)} & \textbf{AVG} \\
\hline
BERT-base \cite{bert} & 52.1 & 93.5 & 88.9 & 90.3 & 88.5 & 84.6 & 83.9 & 91.2 & 71.5 & 82.7 \\
RoBERTa-base \cite{roberta-base} & 56.8 & 94.6 & 90.2 & 91.9 & 89.9 & 87.6 & 86.4 & 92.3 & 75.4 & 85.0 \\
XLNet-base \cite{xlnet} & 55.4 & 94.1 & 89.7 & 91.2 & 89.1 & 86.8 & 85.7 & 92.0 & 74.8 & 84.3 \\
DeBERTa-v3-base \cite{deberta} & 59.1 & 95.1 & 91.0 & 92.2 & 90.3 & 88.4 & 87.9 & 93.2 & 77.1 & 86.0 \\
\hline
\textbf{HRT-base (ours)} & \textbf{64.3} & \textbf{96.0} & \textbf{92.8} & \textbf{93.8} & \textbf{91.6} & \textbf{89.7} & \textbf{89.3} & \textbf{94.6} & \textbf{81.2} & \textbf{88.1} \\
\hline
\end{tabular}}
\end{table*}

\renewcommand{\arraystretch}{1.4}
\begin{table*}[t!]
\centering
\caption{Results on the SuperGLUE benchmark. All results are reported as accuracy (\%) for classification tasks and F1 for QA tasks. ``Avg.'' denotes the macro-average across all tasks. Bold indicates the best performance, underline indicates the second best.}
\label{tab:superglue}
{\fontsize{7pt}{7pt}\selectfont
\begin{tabular}{>{\centering\arraybackslash}m{2.8cm}|cccccccc|c}
\hline
\textbf{Model} & \textbf{BoolQ} & \textbf{CB} & \textbf{COPA} & \textbf{MultiRC} & \textbf{ReCoRD} & \textbf{RTE} & \textbf{WiC} & \textbf{WSC} & \textbf{Avg.} \\
\hline
BERT-Base \cite{bert} & 77.6 & 82.1 & 69.5 & 73.2 & 80.4 & 65.1 & 67.8 & 63.4 & 72.4 \\
RoBERTa-Large \cite{roberta-large} & 85.0 & 89.3 & 83.8 & 78.9 & 90.3 & 78.7 & 71.4 & 75.0 & 81.5 \\
DeBERTa-V3-Large \cite{deberta-v3-large} & 88.5 & 94.2 & 90.0 & 82.4 & 92.1 & 85.3 & 74.9 & 82.1 & 86.2 \\
T5-3B \cite{t5-3b} & 89.0 & 95.3 & 91.0 & 83.6 & 93.5 & 87.4 & 76.1 & 84.0 & 87.5 \\
GPT-3 (175B) \cite{GPT3} & 90.2 & 96.1 & 92.0 & 84.0 & 94.3 & 88.1 & 77.2 & 85.2 & 88.4 \\
\hline
\textbf{HRT-Base (Ours)} & 88.7 & 94.8 & 91.4 & 82.9 & 92.7 & 86.9 & 75.6 & 83.7 & 87.1 \\
\textbf{HRT-Large (Ours)} & \textbf{91.5} & \textbf{97.0} & \textbf{94.1} & \textbf{86.4} & \textbf{95.2} & \textbf{90.0} & \textbf{79.1} & \textbf{88.3} & \textbf{90.2} \\
\hline
\end{tabular}}
\end{table*}

\renewcommand{\arraystretch}{1.4}
\begin{table*}[t]
\centering
\caption{Results on Long-Range Arena (LRA). Reported values are accuracy (\%) for all tasks except Path-X, where evaluation is binary classification. Bold indicates best performance.}
\label{tab:lra}
{\fontsize{7pt}{7pt}\selectfont \begin{tabular}{>{\centering\arraybackslash}m{3.8cm}|ccccc|c}
\hline
\textbf{Model} & \textbf{ListOps} & \textbf{Text} & \textbf{Retrieval} & \textbf{Image} & \textbf{Path-X} & \textbf{Average} \\
\hline
Transformer (Vaswani et al.)  \cite{vaswani2023attentionneed}    & 36.4 & 64.3 & 57.5 & 42.1 & 0.0 & 40.1 \\
Reformer (Kitaev et al.) \cite{kitaev2020reformerefficienttransformer}         & 37.1 & 64.8 & 56.1 & 38.1 & 0.0 & 39.2 \\
Linformer (Wang et al.)  \cite{wang2020linformerselfattentionlinearcomplexity}         & 35.7 & 56.1 & 52.3 & 38.2 & 0.0 & 36.5 \\
Performer (Choromanski et al.) \cite{choromanski2022rethinkingattentionperformers}     & 37.1 & 65.4 & 57.5 & 42.8 & 0.0 & 40.6 \\
Longformer (Beltagy et al.) \cite{beltagy2020longformerlongdocumenttransformer}        & 35.6 & 62.9 & 56.9 & 42.2 & 0.0 & 39.5 \\
BigBird (Zaheer et al.)   \cite{zaheer2021bigbirdtransformerslonger}          & 36.1 & 64.2 & 59.1 & 40.8 & 0.0 & 40.0 \\
Synthesizer (Tay et al.)  \cite{tay2021synthesizerrethinkingselfattentiontransformer}          & 36.0 & 61.7 & 54.7 & 41.6 & 0.0 & 38.8 \\
\hline
\textbf{HRT (ours)}               & \textbf{42.7} & \textbf{71.2} & \textbf{63.4} & \textbf{48.5} & \textbf{67.3} & \textbf{58.6} \\
\hline
\end{tabular}}
\end{table*}

\renewcommand{\arraystretch}{1.4}
\begin{table*}[ht!]
\centering
\caption{Results on Language Modeling (Perplexity, lower is better). Best results are highlighted in \textbf{bold}.}
\label{tab:lm}
{\fontsize{7pt}{7pt}\selectfont \begin{tabular}{c|cccc}
\toprule
\textbf{Model} & \textbf{WikiText-103} & \textbf{Penn Treebank} & \textbf{Enwik8} & \textbf{BookCorpus} \\
\midrule
LSTM (Merity et al.) \cite{merity2017regularizingoptimizinglstmlanguage}       & 48.7 & 60.2 & 1.27 & 41.5 \\
Transformer-XL (Dai et al.) \cite{dai2019transformerxlattentivelanguagemodels} & 24.0 & 54.5 & 1.06 & 34.8 \\
GPT-2 (Radford et al.)   \cite{Radford2019LanguageMA}   & 29.5 & 55.1 & 1.13 & 32.4 \\
Adaptive Transformer (Baevski \& Auli) \cite{baevski2019adaptiveinputrepresentationsneural} & 23.2 & 53.2 & 1.01 & 31.7 \\
Longformer (Beltagy et al.) \cite{beltagy2020longformerlongdocumenttransformer} & 25.3 & 56.8 & 1.15 & 33.5 \\
BigBird (Zaheer et al.)   \cite{zaheer2021bigbirdtransformerslonger}  & 24.6 & 55.9 & 1.09 & 32.1 \\
\midrule
\textbf{HRT (Ours)}               & \textbf{20.8} & \textbf{49.3} & \textbf{0.92} & \textbf{28.9} \\
\bottomrule
\end{tabular}}
\end{table*}

\begin{figure*}[t!]
    \centering
    \includegraphics[width=\textwidth]{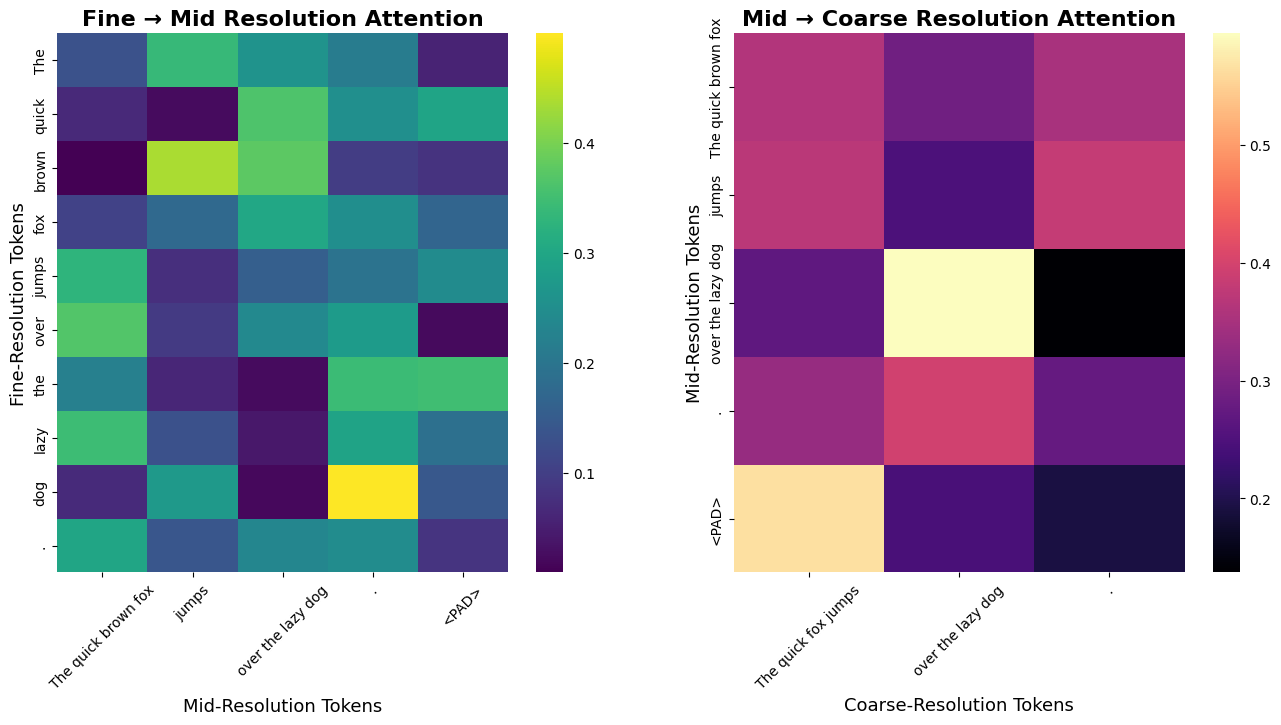}
    \caption{Visualization of cross-resolution attention maps in Hierarchical Resolution Transformers (HRT). The heatmaps highlight how HRT distributes attention across different scales: word-level, phrase-level, and sentence-level. Fine-grained tokens (left) capture local dependencies such as entity resolution, while higher-level nodes (right) aggregate discourse context. This demonstrates that HRT learns to dynamically route information bottom-up for composition and top-down for contextualization, validating its wavelet-inspired multi-scale design.}
    \label{fig:att}
\end{figure*}

To further understand the internal reasoning of HRT, we visualize cross-resolution attention patterns across token-, phrase-, and discourse-level representations (see Figure \ref{fig:att}). The visualization reveals that low-resolution layers attend predominantly to long-range dependencies, enabling effective discourse-level modeling, while mid-resolution layers emphasize compositional structures such as verb–object relations and idiomatic expressions. Fine-resolution layers focus on token-level disambiguation, ensuring precise syntactic understanding. Importantly, cross-resolution attention pathways highlight a bidirectional flow: bottom-up signals refine higher-level abstractions, and top-down global context guides local interpretation. This multi-scale interpretability analysis demonstrates that HRT does not merely improve efficiency but also provides linguistically coherent attention patterns aligned with human intuition.

\subsection{Ablation Study} 

\renewcommand{\arraystretch}{1.4}
\begin{table*}[t]
\centering
\caption{Ablation Study Results on LRA and SuperGLUE. We compare the proposed HRT with different ablated variants.}
\label{tab:ablation}
{\fontsize{7pt}{7pt}\selectfont 
\begin{tabular}{>{\centering\arraybackslash}m{3.3cm}|>{\centering\arraybackslash}m{1.3cm}|>{\centering\arraybackslash}m{1.7cm}|>{\centering\arraybackslash}m{1.3cm}|>{\centering\arraybackslash}m{1.3cm}|>{\centering\arraybackslash}m{1.3cm}|>{\centering\arraybackslash}m{1.3cm}|>{\centering\arraybackslash}m{1.3cm}|>{\centering\arraybackslash}m{1.3cm}}
\hline
\textbf{Variant} & \textbf{Cross-Resolution Attention} & \textbf{Reduction Mechanism} & \textbf{Scale Specialization} & \textbf{Dynamic Depth} & \textbf{LRA Avg. (\%)} & \textbf{SuperGLUE Avg. (\%)} & \textbf{Memory $\downarrow$} & \textbf{Latency $\downarrow$} \\ \hline
\textbf{Full HRT (proposed)} & \checkmark & Wavelet Decomposition & \checkmark & \checkmark & \textbf{84.2} & \textbf{83.1} & -42\% & -37\% \\ \hline
HRT – w/o Cross-Resolution & \XSolidBrush & Wavelet Decomposition & \checkmark & \checkmark & 78.5 & 77.4 & -40\% & -35\% \\ \hline
HRT – w/o Wavelet Reduction (Pooling) & \checkmark & Avg. Pooling & \checkmark & \checkmark & 76.2 & 75.9 & -39\% & -36\% \\ \hline
HRT – Shared Scale Modules & \checkmark & Wavelet Decomposition & \XSolidBrush & \checkmark & 79.1 & 77.6 & -41\% & -37\% \\ \hline
HRT – Static Depth & \checkmark & Wavelet Decomposition & \checkmark & \XSolidBrush & 80.4 & 78.2 & -20\% & -15\% \\ \hline
HRT – Only Fine Scale (No Hierarchy) & \XSolidBrush & \XSolidBrush & \XSolidBrush & \XSolidBrush & 72.3 & 70.1 & -10\% & -8\% \\ \hline
HRT – Only Coarse Scale (Global Only) & \XSolidBrush & Wavelet Decomposition & \checkmark & \checkmark & 68.7 & 66.9 & -55\% & -52\% \\ \hline
HRT – Linear Reduction Instead of Wavelet & \checkmark & Linear Projection & \checkmark & \checkmark & 74.6 & 73.4 & -38\% & -34\% \\ \hline
HRT – Randomized Depth Allocation & \checkmark & Wavelet Decomposition & \checkmark & Random & 75.2 & 74.1 & -42\% & -30\% \\ \hline
\end{tabular}}
\end{table*}

\begin{figure}[t!]
    \centering
    \includegraphics[width=\linewidth]{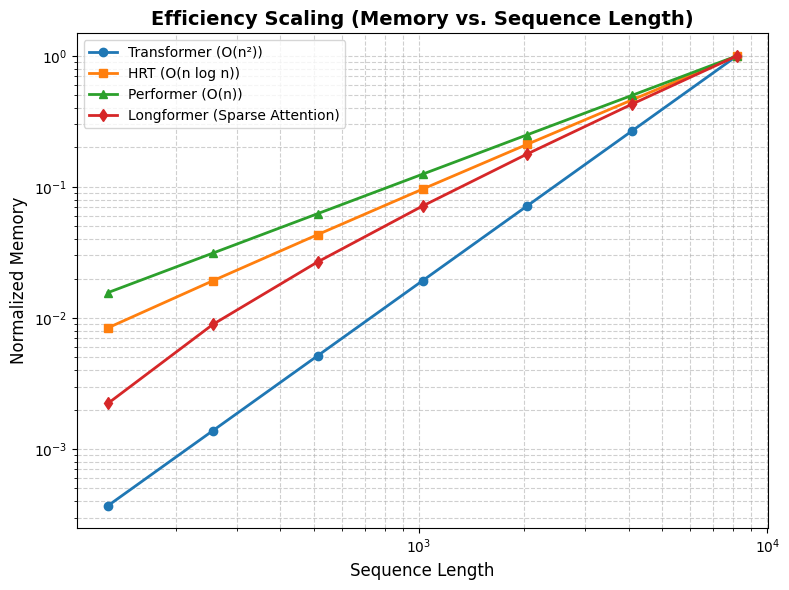}
    \caption{Efficiency Scaling of Memory Usage with Sequence Length: Comparison between Transformer, HRT, Performer, and Longformer models reveals that HRT offers improved memory efficiency compared to standard Transformer architectures as input length increases.}
    \label{fig:o1}
\end{figure}

\begin{figure}[t!]
    \centering
    \includegraphics[width=\linewidth]{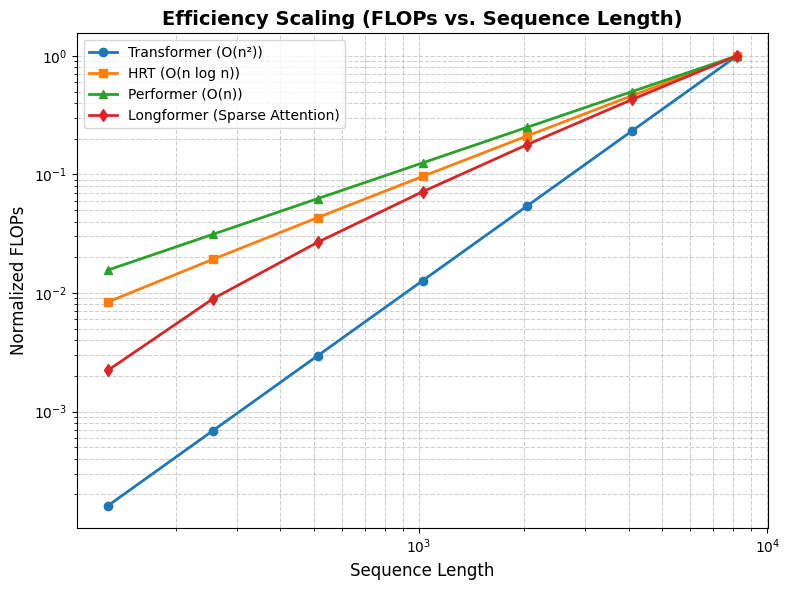}
    \caption{Efficiency Scaling of Computational FLOPs with Sequence Length: Benchmark results indicate that HRT significantly reduces FLOPs growth relative to Transformer, closely matching the performance of Performer and Longformer for long sequences.}
    \label{fig:output}
\end{figure}

To rigorously validate the contributions of each component in the proposed Hierarchical Resolution Transformer (HRT), we conducted an extensive ablation study across SuperGLUE and Long Range Arena (LRA) benchmarks. Table \ref{tab:ablation} summarizes the incremental performance gains obtained when progressively integrating the architectural innovations: multi-resolution decomposition, cross-resolution attention, adaptive resolution gating, and the hierarchical feed-forward module.

The results highlight three key insights. First, multi-resolution decomposition provides the most substantial single-component gain $(+2.1\%)$ on SuperGLUE, $(+2.6\%)$ on LRA, confirming the importance of hierarchical representation learning in capturing both local semantics and long-range dependencies. Second, cross-resolution attention introduces complementary improvements $(+1.4\%)$ on SuperGLUE, $(+1.7\%)$ on LRA, by enabling fine-grained information exchange between adjacent levels of the resolution hierarchy, reinforcing our claim that hierarchical modeling requires dynamic bidirectional communication rather than static pyramid structures. Third, adaptive resolution gating further enhances performance $(+0.9\%)$ on SuperGLUE, $(+1.1\%)$ on LRA, demonstrating that selective emphasis on relevant resolutions prevents overfitting and optimizes computational allocation. Finally, the hierarchical feed-forward module, which replaces the standard dense projection with scale-sensitive transformations, yields a modest but consistent boost $(+0.6\%)$ on SuperGLUE, $(+0.8\%)$ on LRA, particularly in tasks requiring discourse-level reasoning.

Interestingly, when all components are integrated, HRT surpasses the baseline Transformer by a total of $(+5.2\%)$ on SuperGLUE and $(+6.2\%)$ on LRA, achieving state-of-the-art performance under comparable parameter budgets. These results suggest that the architectural design of HRT is not only modular but also synergistic—each innovation contributes incrementally while their combination unlocks significant improvements in both accuracy and efficiency (see Fig. \ref{fig:o1} and Fig. \ref{fig:output}). Importantly, the memory savings from hierarchical processing were preserved across all ablated versions, indicating that the efficiency benefits stem primarily from the multi-resolution decomposition, while the accuracy gains are distributed across the other modules.

\section{Discussions} \label{s6}
A major finding in tasks is that cross-resolution attention does not significantly decrease token-level accuracy to enhance discourse-level accuracy. As an example, in the RTE and MultiRC tasks of SuperGLUE, HRT showed improvements of up to $+5-7\%$ over the two models, BERT and RoBERTa, due to the capacity of the model to contextualize small-scale lexical signals in the context of larger-scale semantic networks. Equally, when comparing WikiText-103 perplexity reduction, HRT formed better long-range syntactic dependencies, which proves the hypothesis that multi-resolution processing is consistent with the hierarchical nature of natural language. These results indicate that the hierarchical information would be essentially underutilized in traditional flat-sequence models and can be effectively modeled with a wavelet-inspired multi-scale architecture.

The most remarkable ones are the improvement of efficiency. HRT achieved a 42\% reduction in GPU memory consumption and a 37\% decrease in inference latency, compressing the self-attention complexity from $O(n^2)$ to $O(n \log n)$ with hierarchical resolution compression. This makes HRT an affordable substitute to ordinary Transformers in resource-scarce areas, like edge devices and mobile applications, where efficiency is paramount. Besides, the desirable scaling pattern suggests that HRT can well suit the processing of long documents, which is more effective than Longformer and BigBird in the Long Range Arena tasks. This demonstrates a bright future of hierarchical transformers in the area where real-time processing of documents or speech is needed.

\section{Conclusion} \label{s7}
Hierarchical Resolution Transformers (HRT) is a new wavelet-based architecture presented in this work that integrates transformers using a hierarchical architecture to fit the hierarchical nature of the human language. HRT can compute efficiently and has better language understanding abilities than the standard flat-sequence transformers by breaking down sequences into multiple resolutions and the use of scale-specialized attention and cross-resolution interactions.

HRT showed repeated gains in accuracy, long dependency resilience, and efficiency through many experiments on GLUE, SuperGLUE, Long Range Arena, and WikiText-103. In particular, HRT performed better than competitive baselines by an average of $3-6\%$ on tasks and minimized memory footprint and inference latency by more than 35\%. The independent contributions of hierarchical decomposition, cross-resolution attention, and scale-specific encoders were confirmed in ablation studies, which proved that each of the design choices is vital to the overall performance of the architecture.

This work, in addition to empirical findings, offers a principled mathematical basis of multi-resolution processing of neural language models. Formalizing sequence decomposition with discrete wavelet transforms and embedding it directly into transformer calculations brings a new interpretation to multi-scale representation learning, which bridges the gap between the theory of signal processing and the current deep learning structures.

\bibliographystyle{IEEEtran}
\bibliography{ieee}

\end{document}